\documentclass[letterpaper]{article} 
\usepackage[preprint]{aaai2027}
\usepackage[hyphens]{url}  
\usepackage{graphicx} 
\usepackage{natbib}  
\usepackage{caption} 
\usepackage{amsmath}
\usepackage{amssymb}
\usepackage{booktabs}
\usepackage{multirow}
\usepackage{tabularx}

\newif\ifshowcomments
\showcommentsfalse

\title{PI-MLLM: Psychophysics-Inspired Multimodal Large Language Model for \\ Oddly Satisfying Video Emotion Analysis}
\author{
    Yuheng Zhang\textsuperscript{1},
    Jie Zhang\textsuperscript{1},
    Yuecong Min\textsuperscript{1},
    Yunwei Zhao\textsuperscript{2}, \\
    Meichen Liu\textsuperscript{2},
    Qiankun Li\textsuperscript{3,4},
    Shiguang Shan\textsuperscript{1}
}
\affiliations{
    \textsuperscript{1}Institute of Computing Technology, Chinese Academy of Sciences\\
    \textsuperscript{2}CNCERT/CC\\
    \textsuperscript{3}Nanyang Technological University\\
    \textsuperscript{4}Imperial College London
}

\begin{document}
\maketitle
\begin{abstract}
Oddly Satisfying Videos (OSVs) can evoke strong viewer responses through repetitive motion, visual tactility, and ASMR-like sounds despite containing few explicit semantic emotion cues. This makes OSVs a challenging setting for Video Emotion Analysis (VEA), since most existing methods focus on semantic video content and under-model the physical sensory dynamics that shape viewer responses. To support Oddly Satisfying Video Emotion Analysis (OSVEA), we construct the Oddly Satisfying Video Dataset (OSVD), comprising $12{,}000$ annotated $5$-second clips. Each clip provides viewer-emotion labels under video-only, audio-only, and audio-visual conditions, together with explanations grounded in visual and acoustic sensory evidence. Building on OSVD, we propose PI-MLLM, a psychophysics-inspired multimodal large language model that integrates Modality-Specific Dynamics Alignment (MSDA) and Coalition-Based Audio-Visual Attribution (CAVA). MSDA uses auxiliary prediction objectives to align modality representations with visual motion and acoustic energy-onset dynamics. CAVA evaluates emotion scores across modality coalitions and applies Shapley attribution to identify the dominant modality and determine whether visual and auditory evidence is synergistic or conflicting. Experiments show that PI-MLLM improves both viewer-side emotion prediction and sensory-grounded explanation quality, suggesting the value of physical sensory cues to video emotion analysis.
\end{abstract}

\section{Introduction}

Understanding how multimedia content shapes human emotion is a central goal in affective computing~\cite{picard1997affective,zhao2019affective,wang2022affective}. Research on video emotion understanding has primarily focused on recognizing emotions expressed by on-screen individuals. A complementary line of research instead predicts emotions elicited in viewers by treating the video itself as an affective stimulus~\cite{dhall2012collecting,zadeh2018mosei,jiang2014predicting,deng2025lcaffect,guo2025stimuvar}. This viewer-centered task, known as Video Emotion Analysis (VEA), is particularly challenging for Oddly Satisfying Videos (OSVs), whose repetitive movements, visual tactility, and ASMR-like sounds can evoke satisfaction or relaxation despite sparse facial, linguistic, and narrative emotion cues~\cite{nansen2022visual,poerio2018more}. Accordingly, Oddly Satisfying Video Emotion Analysis (OSVEA) requires models to move beyond recognizing objects, scenes, and events and capture how physical sensory dynamics shape the viewer's affective experience~\cite{reber2004processing,nansen2022visual}.

\begin{figure*}[!t]
    \centering
    \includegraphics[width=\textwidth,height=0.32\textheight]{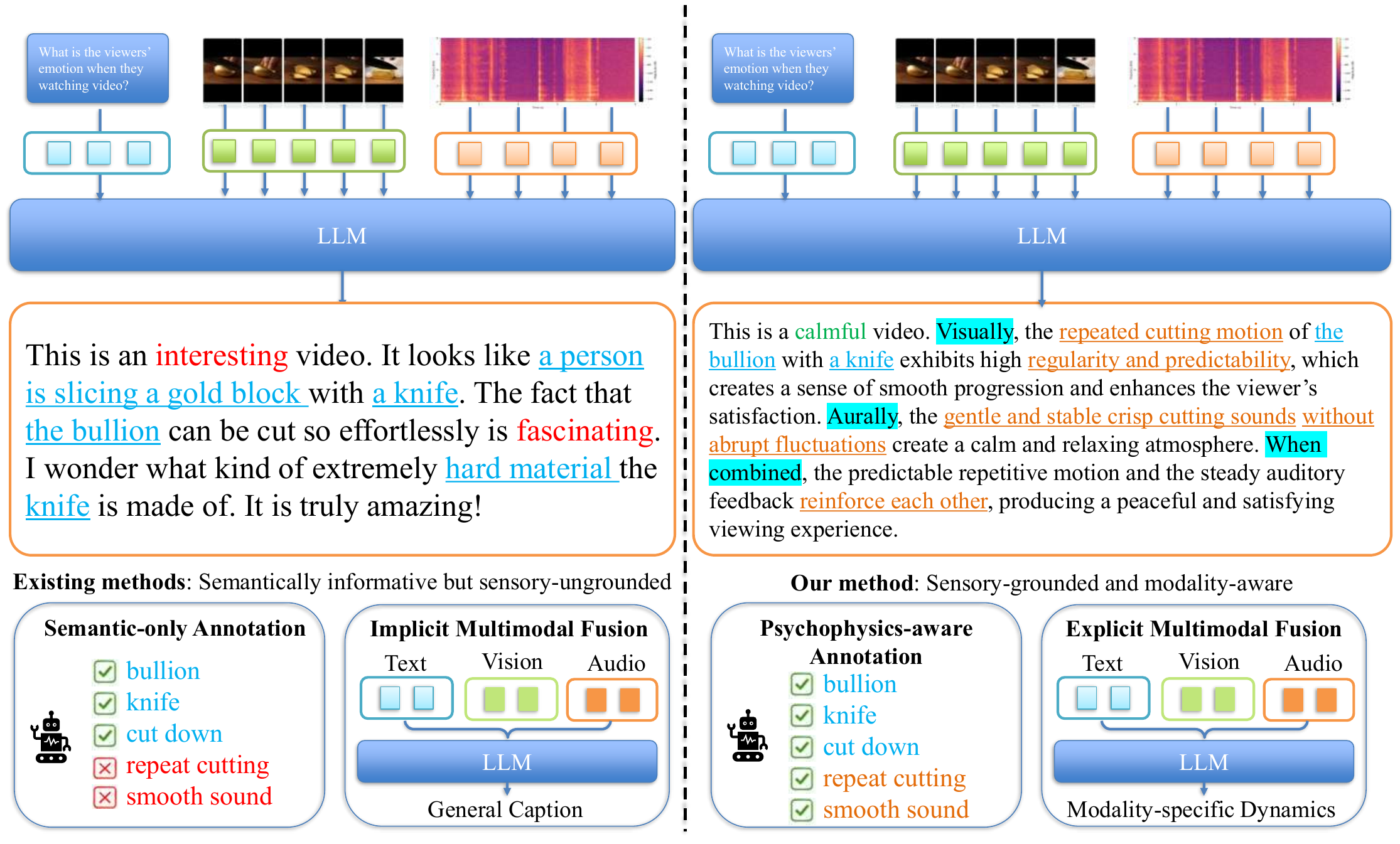}
    \caption{Existing VEA methods rely on sensory-ungrounded semantic annotations and implicit multimodal fusion. Our approach addresses these limitations with psychophysics-aware sensory annotations and explicit multimodal modeling that preserves modality-specific dynamics and cross-modal relations for viewer-emotion reasoning.}
    \label{fig:motivation}
\end{figure*}

However, existing VEA is not well suited to sensory-driven affect modeling in terms of supervision, temporal representation, and multimodal attribution (Figure~\ref{fig:motivation}). Datasets such as EmoCause focus on semantic content and causes, leaving modality-specific sensory dynamics under-explored~\cite{huang2025emodetective}. Although StimuVAR uses optical-flow peaks for frame selection, it may miss the stable repetitive patterns central to OSVs~\cite{guo2025stimuvar}. Likewise, joint-fusion methods such as eMotions obscure modality dominance and whether audio-visual cues reinforce or conflict with each other~\cite{wu2025emotions}. Consequently, a model may recognize a cutting action and predict \emph{interest}, yet fail to explain how predictable motion, rhythmic sounds, and their coordination elicit that response. OSVEA therefore requires modality-specific sensory supervision, dynamics-aware representations, and explicit audio-visual attribution.

To address the lack of an OSVEA-specific dataset, we construct the Oddly Satisfying Video Dataset (OSVD), comprising $12{,}000$ annotated $5$-second clips segmented and selected from $7{,}242$ online OSV source videos. As illustrated in Figure~\ref{fig:annotation_pipeline}, the raw videos are segmented, semantically deduplicated, and filtered for resolution and audio completeness before annotation. Domain-specific visual and audio expert models then process the isolated streams independently: the visual expert annotates the scene, motion, and state change, while the audio expert annotates the sound character, temporal envelope, and spatial impression. A fusion expert reconciles the two reports to generate audio-visual annotations and explicitly records whether the sensory cues reinforce, conflict with or complement each other. Each clip is annotated under video-only, audio-only and audio-visual conditions using the $27$-class Cowen--Keltner taxonomy adopted by VCE~\cite{cowen2017self,mazeika2022how}. Human annotations on a sampled subset are compared with the model-generated labels using Cohen's $\kappa$ to assess reliability.

Building on OSVD, we propose PI-MLLM, a psychophysics-inspired MLLM framework composed of Modality-Specific Dynamics Alignment (MSDA) and Coalition-Based Audio-Visual Attribution (CAVA). Within MSDA, Visual Motion Dynamics (VMD) and Acoustic Energy-Onset Dynamics (AED) align visual and audio representations with signal-derived optical-flow, RMS-loudness, and onset-strength descriptors, encouraging the encoders to preserve sensory dynamics that semantic supervision may overlook. CAVA evaluates emotion scores under different modality coalitions and derives Shapley-style visual and auditory contributions and their interaction~\cite{shapley1953value}, allowing PI-MLLM to distinguish modality dominance, complementarity, and conflict. On OSVD, PI-MLLM achieves an emotion classification accuracy of $43.2\%$ and an explanation semantic similarity of $0.661$, outperforming the evaluated video-only and audio-visual MLLMs as well as the matched autoregressive fine-tuning baseline. Ablations further show that VMD and AED provide complementary gains and that coalition-based attribution yields additional improvements.

The main contributions are summarized as follows:
\begin{itemize}
    \item We introduce OSVD, to our knowledge the first dedicated dataset for OSVEA with fine-grained viewer-side emotion labels and detailed physical sensory-grounded explanations across modalities.
    \item We propose PI-MLLM, a psychophysics-inspired MLLM framework for OSVEA that introduces MSDA to align modality representations with non-semantic visual and acoustic dynamics and CAVA to quantify modality contributions and audio-visual interactions.
    \item Extensive experiments on OSVD demonstrate the effectiveness of PI-MLLM, with comparisons against state-of-the-art methods validating its superior performance.
\end{itemize}

\section{Related Work}

\subsection{Video Emotion Analysis Dataset}
Video emotion analysis dataset targets the emotion triggered in viewers. It is distinguished from traditional multimodal sentiment analysis datasets like MOSI and MOSEI~\cite{zadeh2017tensor,zadeh2018mosei}, which focus on the emotion expressed by the character appeared in the video. Originally, movies were an important material to analyze viewers’ emotion. MediaEval16~\cite{dellandrea2016mediaeval} and LIRIS-ACCEDE~\cite{baveye2015liris} were created to annotate audience feeling after watching specific movie clips. With the development of video platforms, videos are widespread in these platforms for people to share their emotion. YF-E6~\cite{xu2016videoemotion} and VE-8~\cite{jiang2014predicting} were established like mushrooms after rain. They both based on videos from popular video platforms YouTube and Flickr, but annotated based on different emotion classification policy. Recently, VEA datasets come to more detailed annotation. With VCE~\cite{mazeika2022how} annotate the probability distribution of emotion evoked by the video, EmoCause~\cite{huang2025emodetective} annotate emotions with abundant explanations including video,audio, shot and content analysis. However, they still lack the annotation of some physical sensory features and the modality-specific emotion labels.
In this work, we introduce OSVD, a dataset with unimodal and cross-modal sensory annotation and detailed explanation of physical sensory features which also dominant to viewer’s emotion.

\subsection{Video Emotion Analysis Methods}
The goal of VEA is to extract the relevant features to bridge the gap between the video content and the triggered emotion. Early studies on VEA mainly center on designing hand-crafted features to represent emotion. For example, Acar et al. utilized Mel-frequency cepstral coefficients to capture speech features relevant to viewer’s emotion~\cite{acar2017comprehensive}. However, the correlation between the simple feature and triggered emotion is hard to be modeled. With the emergence of various VEA dataset, methods begun to focus on the multi-modal information fusion. Representative works like TAM~\cite{pan2022tam} designed a cross-temporal multimodal fusion module that applies attention-based fusion to different modalities within and across video segments in order to capture the cross-modal feature central to emotion analysis. However, the fusion process may lack intuitive explanation and thus affecting the credibility of the results. Recent work begun to treat external information as additional cues for emotion analysis. For example, LCAffect~\cite{deng2025lcaffect} utilize Danmaku as an auxiliary modality for emotion analysis. EmoDETective~\cite{huang2025emodetective} utilize detailed annotation of audio visual content and shot description to improve emotion analysis explainability. However, these methods may lack relative information in many situations.
Motivated by predictive-coding and processing-fluency accounts, we use optical-flow dynamics as a descriptor of visual motion predictability relevant to affective responses~\cite{friston2012predictive,reber2004processing}.

\section{OSVD Construction}
\label{sec:osvd}

\begin{figure*}[!t]
    \centering
    \includegraphics[width=\textwidth]{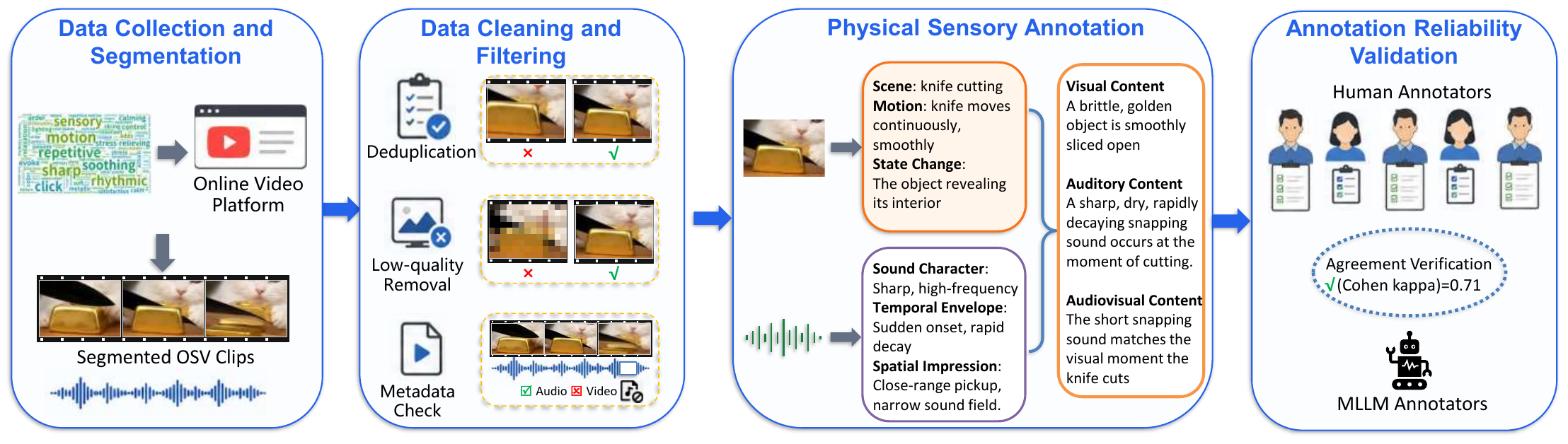}
    \caption{\textbf{OSVD construction and annotation pipeline.} Word-cloud-derived keywords guide data collection. Then segmented clips are semantically deduplicated, filtered for resolution, and checked for audio integrity. Visual annotations are structured as scene, motion, and state change, while audio annotations are structured as sound character, temporal envelope, and spatial impression. Finally, the modality-specific reports are fused into audio-visual annotations and evaluated against human judgments.}
    \label{fig:annotation_pipeline}
\end{figure*}

To support OSVEA, we construct the Oddly Satisfying Video Dataset (OSVD), with $12{,}000$ annotated $5$-second clips segmented and selected from $7{,}242$ online short-form OSV source videos. As shown in Table~\ref{tab:dataset_compare}, existing VEA datasets typically annotate only one viewing condition and, at most, provide semantic explanations. OSVD instead provides viewer-emotion labels under audio, visual, and audio-visual conditions, together with both semantic and physical sensory explanations. Figure~\ref{fig:annotation_pipeline} summarizes its construction and annotation pipeline.

\begin{table}[!t]
\centering
\caption{\textbf{Comparison between OSVD and representative VEA datasets.} Emotion-label columns indicate whether annotations are provided under audio-only (A), video-only (V), and audio-visual (AV) viewing conditions. Emotion explanations are divided into semantic descriptions and physical sensory descriptions.}
\label{tab:dataset_compare}
\scriptsize
\setlength{\tabcolsep}{2.0pt}
\renewcommand{\arraystretch}{1.05}
\resizebox{\columnwidth}{!}{%
\begin{tabular}{@{}lcccccc@{}}
\toprule
\multirow{2}{*}{Dataset} & \multirow{2}{*}{Scale} & \multicolumn{3}{c}{\textbf{Emotion Label}} & \multicolumn{2}{c}{\textbf{Emotion Explanation}} \\
\cmidrule(lr){3-5}\cmidrule(l){6-7}
 & & A & V & AV & Semantic & \begin{tabular}{@{}c@{}}Physical\\sensory\end{tabular} \\
\midrule
LIRIS-ACCEDE~\cite{baveye2015liris} & 9,800 & $\times$ & $\times$ & $\checkmark$ & $\times$ & $\times$ \\
VE-8~\cite{jiang2014predicting} & 1,101 & $\times$ & $\times$ & $\checkmark$ & $\times$ & $\times$ \\
YF-E6~\cite{xu2016videoemotion} & 1,637 & $\times$ & $\times$ & $\checkmark$ & $\times$ & $\times$ \\
VAD~\cite{wang2024vad} & 19,267 & $\times$ & $\times$ & $\checkmark$ & $\times$ & $\times$ \\
VCE~\cite{mazeika2022how} & 61,046 & $\times$ & $\checkmark$ & $\times$ & $\times$ & $\times$ \\
EmoCause~\cite{huang2025emodetective} & 14,000 & $\times$ & $\times$ & $\checkmark$ & $\checkmark$ & $\times$ \\
OSVD (ours) & 12,000 & $\checkmark$ & $\checkmark$ & $\checkmark$ & $\checkmark$ & $\checkmark$ \\
\bottomrule
\end{tabular}}
\end{table}

\paragraph{Collection and segmentation.}
We derive OSV-related search terms from a word cloud of representative descriptions and use them to retrieve raw videos from online video platforms. Each collected video is standardized and segmented into two non-overlapping $5$-second clips. OSVs are often organized around short repeated actions such as slicing, crushing, pouring, cleaning, or brushing~\cite{nansen2022visual,gallagher2016eliciting}; a $5$-second clip usually captures one complete micro-action sufficient for emotion analysis.

\paragraph{Data cleaning and filtering.}
We conduct several cleaning and filtering steps to ensure data quality. Firstly, we calculate pairwise semantic similarity between candidate videos and remove one video from each pair whose score exceeds $0.9$. Secondly, we discard videos whose orientation-normalized resolution is below $360\times640$ pixels. Thirdly, we inspect media-container and stream metadata, removing data with missing or corrupted audio.

\paragraph{Emotion and physical sensory annotation.}
The visual and audio experts independently annotate each clip under isolated unimodal conditions. For the visual stream, \textit{scene} describes the semantic setting and activity, \textit{motion} captures time-varying movement patterns, and \textit{state change} records observable transformations caused by the activity. With only video stream provided, Molmo2~\cite{clark2026molmo2} uses these information to produce a video-only emotion label and explanation. For the audio stream, \textit{sound character} describes timbral and spectral qualities, \textit{temporal envelope} captures onset, evolution, and decay, and \textit{spatial impression} represents perceived source proximity and sound-field extent. Given only this stream, Music Flamingo~\cite{ghosh2026musicflamingo} uses the three fields to produce an audio-only label and explanation. Both labels use the same fixed $27$-class viewer-emotion taxonomy. Claude Sonnet 4.6~\cite{anthropic2026sonnet46} integrates the visual and auditory content reports to generate the audio-visual label and explanation, describing whether the two streams reinforce, conflict with, or complement each other.

\paragraph{Annotation reliability validation.}
To assess whether the expert-model labels reflect human viewer judgments, we randomly sample $100$ clips from OSVD and ask five independent annotators to assign emotion labels for each modality view. Table~\ref{tab:human} reports the average Cohen's $\kappa$ between the annotators and the corresponding OSVD labels under the expert models' view. The values show that OSVD annotations are consistently aligned with human judgments across modalities.

\begin{table}[!t]
\centering
\caption{Average human agreement with OSVD labels on $100$ randomly sampled clips under three modality views.}
\label{tab:human}
\resizebox{\columnwidth}{!}{%
\begin{tabular}{lccc}
\toprule
Metric & Video-only & Audio-only & Audio-visual \\
\midrule
Avg. Cohen's $\kappa$ & 0.70 & 0.67 & 0.71 \\
\bottomrule
\end{tabular}}
\end{table}

\section{Method}
\label{sec:method}

\subsection{Overview}

\begin{figure*}[!t]
    \centering
    \includegraphics[width=0.94\linewidth]{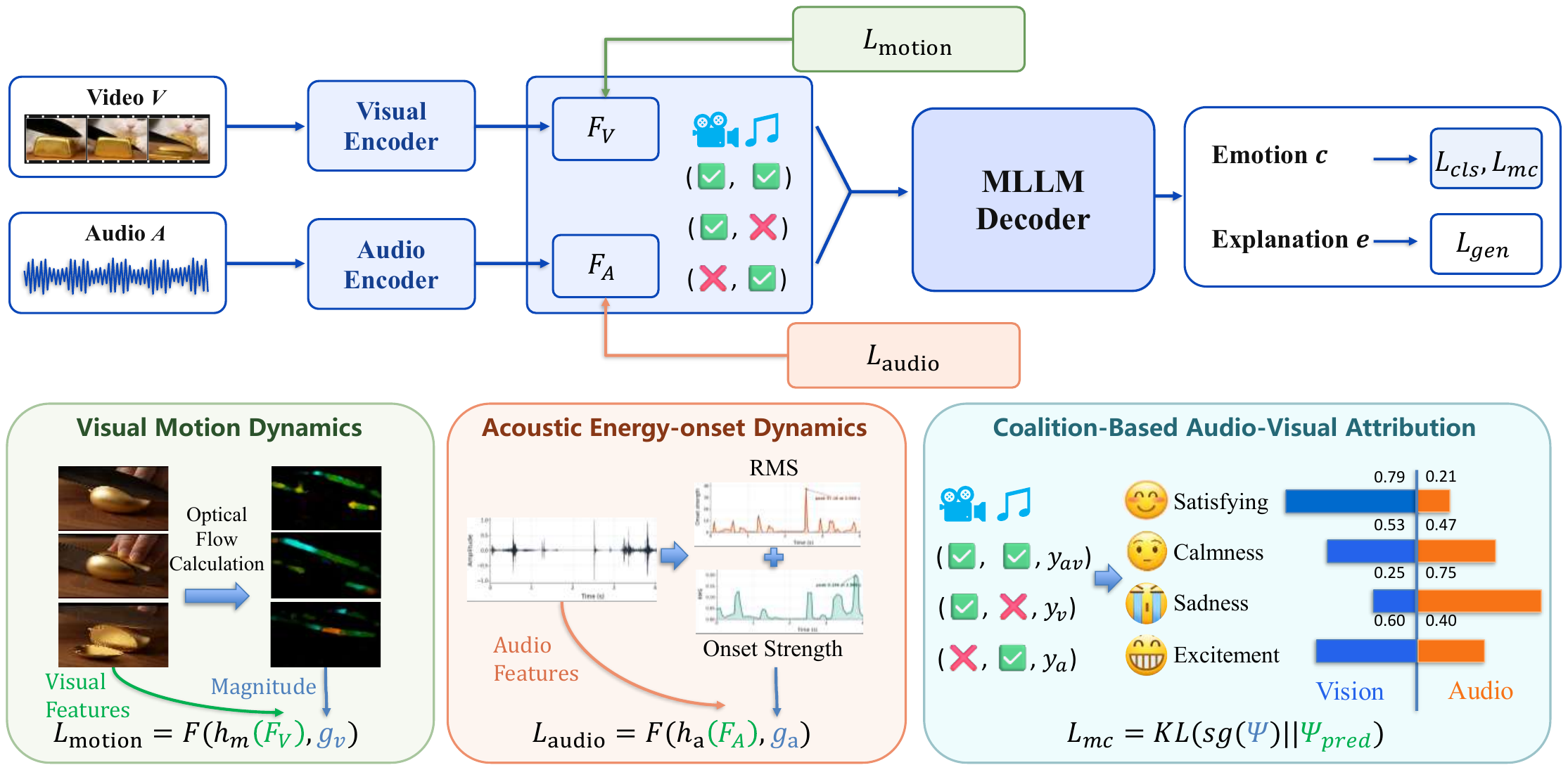}
    \caption{\textbf{Overall framework of PI-MLLM.} Visual and audio inputs are encoded into $F_V$ and $F_A$ to generate viewer emotion $c$ and a sensory-grounded explanation $e$. MSDA aligns the visual representation with optical-flow dynamics and the audio representation with RMS loudness and onset strength. CAVA uses modality-conditioned emotion scores to estimate visual and auditory contributions and their interaction. The total objective combines generation, dynamics-alignment, coalition-classification, and attribution losses.}
    \label{fig:framework}
\end{figure*}

Given an audio-visual clip $x=(V,A)$, PI-MLLM predicts $y=(c,e)$, where $c$ is the viewer-side emotion label and $e$ is its sensory-grounded explanation. As illustrated in Figure~\ref{fig:framework}, the visual and audio encoders produce features $F_V$ and $F_A$ for the MLLM decoder to generate $c$ and $e$. PI-MLLM contains two complementary modules: Modality-Specific Dynamics Alignment (MSDA) and Coalition-Based Audio-Visual Attribution (CAVA). We describe the two modules below.

\subsection{Modality-Specific Dynamics Alignment}
\label{subsec:sensory_supervision}

\paragraph{Visual Motion Dynamics (VMD).}
Prediction-error accounts suggest that aesthetic pleasure can arise when prediction errors become progressively reduced as perceptual regularities are learned~\cite{friston2012predictive,reber2004processing}. Taking a metal-cutting video as an example, after observing repeated cuts, viewers can infer the knife-motion pattern and anticipate subsequent changes. The resulting reduction in prediction error may contribute to the satisfying response. Motivated by this theory, we use mean optical-flow magnitude to represent frame-to-frame visual discrepancy and guide the model to better capture motion predictability. We attach a lightweight motion-regression head $h_m$ to the visual encoder to predict mean optical-flow magnitude in order to help capture visual motion dynamics which are othervise neglected by previous semantic training strategy. Each clip is divided into $T$ temporal windows. At step $t$, we compute the mean Farneback optical-flow magnitude $g_{v,t}\in\mathbb{R}$ over consecutive frame pairs in the corresponding window~\cite{farneback2003two}. We supervise the time-aligned visual features with a mean-squared error loss:

\begin{equation}
    \mathcal{L}_{\text{motion}} = \frac{1}{T}\sum_{t=1}^{T}
    \lVert h_m(F_V)_t - g_{v,t} \rVert_2^2 .
\end{equation}
This objective encourages the visual representation to retain visual motion dynamics that may otherwise be weakly exposed to the text decoder for emotion prediction.

\paragraph{Acoustic energy-onset dynamics (AED).}
Poerio et al. report that ASMR experiences combine overall calmness with localized physiological activation~\cite{poerio2018more}. This coexistence suggests that affective audio should be characterized by both its sustained energy profile and its brief local features. Accordingly, we use frame-wise root-mean-square (RMS) loudness to describe the overall acoustic-energy trajectory relevant to perceived calmness, and onset strength to capture localized acoustic transients relevant to brief activation. At the same $T$ time steps, these two complementary descriptors form the target $g_{a,t}\in\mathbb{R}^{2}$. A single regression head $h_a$ is attached to the audio encoder to predict the descriptors from the audio hidden features $F_A$, and we also supervise it with a mean-squared error loss:

\begin{equation}
    \mathcal{L}_{\text{audio}} = \frac{1}{T}\sum_{t=1}^{T} \lVert h_a(F_A)_t - g_{a,t} \rVert_2^2 .
\end{equation}
By reconstructing both descriptors, AED encourages the audio representation to preserve the non-semantic acoustic dynamics that distinguish sustained calmness from localized activation. These dynamics also provide complementary sensory evidence for predicting and explaining viewer-side emotion.

\subsection{Coalition-Based Audio-Visual Attribution}
\label{subsec:av_module}

Audio and vision do not always contribute to OSV affect consistently. A sharp cutting sound may reduce the relaxation suggested by smooth visual motion, while a soft repetitive sound may strengthen a visually calm ASMR clip. Shapley values provide a coalition-based attribution of model outputs~\cite{shapley1953value}. PaSE incorporates Shapley-based modality equilibrium into multimodal sentiment training~\cite{he2026pase}. PI-MLLM instead requires modality contributions and interactions to represent genuine dominance, conflict, or complementarity rather than forcing the modalities to balance. CAVA therefore considers two sensory modalities, $V$ and $A$, and four coalitions $S\in\{\emptyset,\{V\},\{A\},\{V,A\}\}$, whose emotion scores are used to derive both Shapley contributions and pairwise interaction.

Specifically, we use a shared auxiliary classifier $q_{\psi}$ over the $27$ emotion classes to predict the emotion labels. For each coalition $S$, the coalition classifier is written as:
\begin{equation}
    q_{\psi}(c\mid S) = \mathrm{softmax}\big(f_{\psi}(S)\big)_c,
\end{equation}
where $\psi$ denotes the classifier parameters, $f_{\psi}(S)$ denotes the $27$-class logits computed from the available modality features in $S$, and $q_{\psi}(\cdot\mid S)$ is the resulting coalition-level emotion distribution. OSVD provides three supervised coalition labels: the video-only emotion label $y_v$, the audio-only emotion label $y_a$, and the fused audio-visual emotion label $y_{av}$. We train the head with:
\begin{equation}
\begin{aligned}
    \mathcal{L}_{\text{cls}} ={}&
    \mathrm{CE}\!\left(q_{\psi}(\cdot\mid\{V\}), y_v\right)
    + \mathrm{CE}\!\left(q_{\psi}(\cdot\mid\{A\}), y_a\right) \\
    &+ \mathrm{CE}\!\left(q_{\psi}(\cdot\mid\{V,A\}), y_{av}\right).
\end{aligned}
\end{equation}

The coalition-level score for the final label $y_{av}$ is defined in log-odds space:
\begin{equation}
    s_y(S)=
    \log \frac{\tilde{q}_{\psi}(y_{av}\mid S)}
    {1-\tilde{q}_{\psi}(y_{av}\mid S)},
\end{equation}
where $\tilde{q}_{\psi}(y_{av}\mid S)=\mathrm{clip}(q_{\psi}(y_{av}\mid S),\delta,1-\delta)$, and $\delta$ is a small clipping constant for numerical stability. 

\paragraph{Shapley contribution computation.}
We  compute the Shapley contributions of vision and audio:
\begin{equation}
\begin{aligned}
    \phi_V ={}& \frac{1}{2}\big[s_y(\{V\})-s_y(\emptyset)\big]
    + \frac{1}{2}\big[s_y(\{V,A\})-s_y(\{A\})\big], \\
    \phi_A ={}& \frac{1}{2}\big[s_y(\{A\})-s_y(\emptyset)\big]
    + \frac{1}{2}\big[s_y(\{V,A\})-s_y(\{V\})\big].
\end{aligned}
\end{equation}
Here, a positive value means the modality increases the log-odds score of the final viewer-side emotion. To make the attribution signal learnable during training, we further introduce a modality contribution loss. We directly use the signed Shapley contributions $\phi_V$ and $\phi_A$ as the visual and auditory contribution scores, and normalize their magnitudes into a two-way contribution target:
\begin{equation}
    \boldsymbol{\psi} =
    \left[
    \frac{|\phi_V|}{|\phi_V|+|\phi_A|+\epsilon},
    \frac{|\phi_A|}{|\phi_V|+|\phi_A|+\epsilon}
    \right],
\end{equation}
where $\epsilon$ is a small constant. A contribution head predicts a distribution $\boldsymbol{\psi}_{\text{pred}}$ over the visual and auditory modalities, and we supervise it with:
\begin{equation}
    \mathcal{L}_{\text{mod-contrib}} =
    \mathrm{KL}\!\left(\mathrm{stopgrad}(\boldsymbol{\psi})\,\Vert\,\boldsymbol{\psi}_{\text{pred}}\right).
\end{equation}
This modality contribution loss teaches the model which modality contributes more to the final viewer-side emotion.

\paragraph{Shapley interaction computation.}
We further define the pairwise audio-visual interaction as:
\begin{equation}
    I_{VA}=s_y(\{V,A\})-s_y(\{V\})-s_y(\{A\})+s_y(\emptyset).
\end{equation}
A positive $I_{VA}$ indicates synergy beyond the sum of unimodal evidence, while a negative $I_{VA}$ indicates conflict or redundancy. This formulation provides an interpretable estimate of modality interaction rather than forcing the two modalities to agree.We also normalize the interaction effect as:
\begin{equation}
    \hat{I}_{VA} =
    \frac{I_{VA}}{|\phi_V|+|\phi_A|+|I_{VA}|+\epsilon},
\end{equation}
and train an interaction head to predict $I_{\text{pred}}$:
\begin{equation}
    \mathcal{L}_{\text{mod-int}} =
    \mathrm{Huber}\!\left(I_{\text{pred}}, \mathrm{stopgrad}(\hat{I}_{VA})\right).
\end{equation}
This modality interaction loss keeps the interaction target robust when one coalition score is noisy, while the sign of $\hat{I}_{VA}$ still indicates whether audio and vision form a positive or negative interaction. The combined-modality loss is denoted as:
\begin{equation}
    \mathcal{L}_{\text{cm}} =
    \mathcal{L}_{\text{mod-contrib}} + \mathcal{L}_{\text{mod-int}} .
\end{equation}

\subsection{Joint Training Objective}
\label{subsec:training_objective}

Combining the autoregressive generation loss $\mathcal{L}_{\text{gen}}$ with the auxiliary losses from sessions above, the total training objective is:
\begin{equation}
\begin{aligned}
    \mathcal{L} =
    &\mathcal{L}_{\text{gen}}
    + \lambda_1 \mathcal{L}_{\text{motion}}
    + \lambda_2 \mathcal{L}_{\text{audio}}
    + \lambda_3 \mathcal{L}_{\text{cls}} \\
    &+ \lambda_4 \mathcal{L}_{\text{cm}},
\end{aligned}
\end{equation}
where $\mathcal{L}_{\text{gen}}$, $\mathcal{L}_{\text{motion}}$, $\mathcal{L}_{\text{audio}}$, $\mathcal{L}_{\text{cls}}$, and $\mathcal{L}_{\text{cm}}$ denote the autoregressive generation, motion, audio, classification, and combined-modality losses respectively. The weights $\lambda_{1},\ldots,\lambda_{4}$ balance the contribution of each component and are set to zero when the corresponding module is ablated.

\section{Experiments}
\label{sec:experiments}

\subsection{Experimental Settings}
\label{subsec:settings}

\noindent\textbf{Dataset and protocol.}
The primary evaluation is on OSVD because it provides OSV-specific, modality-aware viewer-side labels and sensory-grounded explanations. We use a class-balanced OSVD test set to prevent majority emotions from dominating the metric. Emotion classification is evaluated using top-1 accuracy (ACC\%) over the 27-class Cowen--Keltner taxonomy, while explanation quality is measured using Sentence-BERT-based Reason Similarity (RS)~\cite{reimers2019sentencebert}. We further report an out-of-distribution protocol on VCE and VE-8 after converting their labels to the OSVD taxonomy.

\noindent\textbf{Backbone and implementation.}
We instantiate the backbone with Qwen3-Omni~\cite{xu2025qwen3omni} and fine-tune it with LoRA~\cite{hu2022lora} ($r{=}64$).

\noindent\textbf{Baselines.}
External baselines span four groups: video MLLMs (VideoLLaMA3~\cite{zhang2025videollama3}, LLaVA-Video~\cite{zhang2024llavavideo}, Qwen2.5-VL~\cite{bai2025qwen25vl}), audio-visual MLLMs (Qwen2.5-Omni~\cite{Qwen2.5-Omni}, video-SALMONN-2+~\cite{tang2025videosalmonn2}), a closed commercial model (Claude Sonnet 4.6~\cite{anthropic2026sonnet46}) and video emotion analysis methods (Emotion-LLaMA~\cite{cheng2024emotionllama}, StimuVAR~\cite{guo2025stimuvar}, EmoDETective~\cite{huang2025emodetective}). We additionally report Qwen3-Omni ~\cite{xu2025qwen3omni} as the matched baseline for autoregressive(AR) fine-tuning and PI-MLLM.

\subsection{Comparison with SOTA Methods}
\label{subsec:main_results}

\begin{table}[!t]
\centering
\caption{\textbf{Comparison with SOTA methods on OSVD.} ACC (\%) denotes $27$-class viewer-side emotion classification accuracy, and RS denotes semantic similarity between generated and reference explanations. Best per column in \textbf{bold}.}
\label{tab:sota_comparison}
\footnotesize
\setlength{\tabcolsep}{1.75pt}
\renewcommand{\arraystretch}{0.96}
\begin{tabular}{@{}lcc@{}}
\toprule
Method & ACC (\%) & RS \\
\midrule
\multicolumn{3}{l}{\textit{Video MLLMs}} \\
VideoLLaMA3-7B~\cite{zhang2025videollama3}    & 11.8 & 0.322 \\
LLaVA-Video-7B~\cite{zhang2024llavavideo}     & 14.6 & 0.487 \\
Qwen2.5-VL-7B~\cite{bai2025qwen25vl}          & 23.8 & 0.472 \\
\midrule
\multicolumn{3}{l}{\textit{Audio-visual MLLMs}} \\
Qwen2.5-Omni-7B~\cite{Qwen2.5-Omni}           & 17.6 & 0.498 \\
video-SALMONN-2+~\cite{tang2025videosalmonn2} & 18.5 & 0.463 \\
\midrule
\multicolumn{3}{l}{\textit{Closed commercial model}} \\
Claude Sonnet 4.6~\cite{anthropic2026sonnet46}  & 32.6 & 0.508 \\
\midrule
\multicolumn{3}{l}{\textit{Video emotion analysis methods}} \\
Emotion-LLaMA-7B~\cite{cheng2024emotionllama}  & 11.3 & 0.301 \\
StimuVAR-7B~\cite{guo2025stimuvar}             & 9.1 & 0.461 \\
\midrule
\multicolumn{3}{l}{\textit{Qwen3-Omni baseline and our methods}} \\
Qwen3-Omni-30B (base)~\cite{xu2025qwen3omni}  & 29.2 & 0.535 \\
Qwen3-Omni-30B + AR fine-tuning & 35.1 & 0.636 \\
\textbf{Qwen3-Omni-30B + PI-MLLM}              & \textbf{43.2} & \textbf{0.661} \\
\bottomrule
\end{tabular}
\end{table}

As shown in Table~\ref{tab:sota_comparison}, we report the classification accuracy and semantic similarity of different methods on OSVD. Overall, PI-MLLM achieves the best performance, with 43.2\% accuracy and 0.661 semantic similarity.

For vision-only models, Qwen2.5-VL and LLaVA-Video achieve the best results in accuracy (23.8\%) and semantic similarity (0.487), respectively. PI-MLLM outperforms them by 19.4\% and 0.174, showing the benefit of incorporating auditory information.

Audio-visual models remain limited, with video-SALMONN-2+ reaching 18.5\% accuracy and Qwen2.5-Omni achieving 0.498 similarity. PI-MLLM still improves over them by 24.7\% and 0.163, indicating that simple multimodal fusion is insufficient for OSVD.

Video emotion analysis methods perform even worse, where Emotion-LLaMA and StimuVAR achieve 11.3\% accuracy and 0.461 similarity, respectively. PI-MLLM surpasses them by 31.9\% and 0.200, highlighting the difficulty of modeling OSVD with conventional approaches.

Among closed-source baselines, Claude Sonnet 4.6 is the strongest, with 32.6\% accuracy and 0.508 similarity. PI-MLLM still leads by 10.6\% and 0.153.

Overall, these results demonstrate that PI-MLLM effectively improves viewer emotion understanding in OSVD by leveraging modality-specific sensory dynamics and cross-modal interactions.

\subsection{Module Ablation}
\label{subsec:ablation}

\begin{table*}[!t]
\centering
\caption{\textbf{Ablation study on OSVD.} VMD and AED denote the visual motion dynamics and acoustic energy-onset dynamics auxiliary objectives inside MSDA, respectively. CLS, MC, and MI denote the coalition emotion classification loss, modality contribution loss, and modality interaction loss inside CAVA, respectively.}
\label{tab:ablation_full}
\resizebox{\textwidth}{!}{%
\begin{tabular}{lcccccccc}
\toprule
Configuration & VMD & AED & CLS & MC & MI & ACC (\%) & RS & $\Delta$ ACC (\%) \\
\midrule
Qwen3-Omni + AR~\cite{xu2025qwen3omni} & -- & -- & -- & -- & -- & 35.1 & 0.636 & -- \\
AR + VMD & $\checkmark$ & -- & -- & -- & -- & 38.8 & 0.642 & +3.7 \\
AR + AED & -- & $\checkmark$ & -- & -- & -- & 39.7 & 0.646 & +4.6 \\
AR + VMD + AED & $\checkmark$ & $\checkmark$ & -- & -- & -- & 42.1 & 0.655 & +7.0 \\
AR + VMD + AED + MC & $\checkmark$ & $\checkmark$ & -- & $\checkmark$ & -- & 42.7 & 0.659 & +7.6 \\
AR + VMD + AED + CLS + MC & $\checkmark$ & $\checkmark$ & $\checkmark$ & $\checkmark$ & -- & 42.8 & 0.659 & +7.7 \\
\textbf{PI-MLLM} & $\checkmark$ & $\checkmark$ & $\checkmark$ & $\checkmark$ & $\checkmark$ & \textbf{43.2} & \textbf{0.661} & \textbf{+8.1} \\
\bottomrule
\end{tabular}}
\end{table*}

Table~\ref{tab:ablation_full} presents the contribution of each component in PI-MLLM. Following the Qwen3-Omni autoregressive baseline, we progressively introduce the proposed supervision objectives to evaluate their individual and combined effects.

MSDA consistently improves performance. Introducing the visual motion dynamics objective alone increases the classification accuracy from 35.1\% to 38.8\%, while the acoustic energy-onset dynamics objective achieves a larger improvement of 39.7\%. Combining both objectives further raises the classification accuracy to 42.1\% and the semantic similarity to 0.655, demonstrating that visual motion and acoustic dynamics provide complementary sensory cues for viewer emotion understanding.

We then evaluate the CAVA components. Based on the model equipped with both MSDA objectives, adding modality contribution supervision improves the classification accuracy from 42.1\% to 42.7\%. Introducing coalition emotion classification further increases the accuracy to 42.8\%. Finally, incorporating modality interaction supervision yields the full PI-MLLM, achieving the best performance with 43.2\% classification accuracy and 0.661 semantic similarity.

Overall, every proposed component consistently contributes to the final performance. The MSDA objectives enable the model to capture visual and auditory physical cues, while the CAVA objectives explicitly supervise coalition emotion prediction, modality contribution estimation, and cross-modal interaction modeling. Their complementary effects lead to the strongest overall performance on OSVD.

\subsection{Out-of-Distribution Generalization}
\label{subsec:cross}

\begin{table}[!t]
\centering
\caption{\textbf{Out-of-distribution evaluation on VCE and VE-8.} Both datasets are evaluated after re-annotation under the OSVD viewer-side $27$-class taxonomy and sensory-explanation protocol.}
\label{tab:cross}
\scriptsize
\setlength{\tabcolsep}{1.4pt}
\renewcommand{\arraystretch}{1.05}
\begin{tabular}{lcccc}
\toprule
Method & \multicolumn{2}{c}{VCE} & \multicolumn{2}{c}{VE-8} \\
\cmidrule(lr){2-3}\cmidrule(lr){4-5}
& ACC (\%) & RS & ACC (\%) & RS \\
\midrule
Qwen3-Omni~\cite{xu2025qwen3omni} & 27.6 & 0.486 & 29.1 & 0.501 \\
\midrule
\multicolumn{5}{l}{\textit{Video emotion analysis methods}} \\
Emotion-LLaMA~\cite{cheng2024emotionllama} & 22.4 & 0.368 & 24.7 & 0.383 \\
StimuVAR~\cite{guo2025stimuvar} & \textbf{34.3} & 0.505 & 35.5 & 0.471 \\
EmoDETective~\cite{huang2025emodetective} & 31.9 & 0.497 & 31.4 & 0.518 \\
\midrule
\multicolumn{5}{l}{\textit{Our methods}} \\
Qwen3-Omni + AR finetuning & 30.9 & 0.524 & 32.6 & 0.536 \\
Qwen3-Omni + PI-MLLM & 33.8 & \textbf{0.552} & \textbf{36.1} & \textbf{0.567} \\
\bottomrule
\end{tabular}
\end{table}

Table~\ref{tab:cross} evaluates the out-of-distribution generalization ability of PI-MLLM on VCE and VE-8. Since neither dataset originally follows the OSVD annotation protocol, both are re-annotated using the same 27-class viewer-side emotion taxonomy and sensory-grounded explanation format to enable a fair comparison.

As expected, all methods perform worse than on OSVD because of the domain shift in video content, emotion distributions, and annotation styles. Nevertheless, PI-MLLM demonstrates strong generalization across both datasets. On VCE, StimuVAR achieves the highest classification accuracy (34.3\%), while PI-MLLM obtains a comparable classification accuracy of 33.8\% and the highest semantic similarity score of 0.552. On VE-8, PI-MLLM achieves the best performance on both metrics, reaching 36.1\% classification accuracy and 0.567 semantic similarity. Compared with the strongest competing methods, PI-MLLM improves the semantic similarity by 0.047 on VCE and improves the classification accuracy and semantic similarity on VE-8 by 0.6 percentage points and 0.049, respectively.

Compared with the autoregressive fine-tuning baseline, PI-MLLM consistently improves both classification accuracy and semantic similarity on the two datasets, demonstrating that the proposed supervision generalizes beyond the OSVD training distribution. Overall, these results indicate that explicitly modeling modality-specific sensory dynamics together with modality contributions and interactions produces representations that transfer effectively to unseen video emotion datasets.

\section{Conclusion}
This work investigates viewer emotion understanding in Oddly Satisfying Videos and highlights the importance of audio-visual sensory dynamics that are often overlooked by semantic-centered VEA methods. Through OSVD and PI-MLLM, we incorporated modality-aware emotion supervision, modality-specific dynamics alignment, and coalition-based audio-visual attribution into a unified framework for OSVEA. Experimental results demonstrate that modeling these factors improves both viewer-emotion prediction and sensory-grounded explanation, while the ablation studies confirm the complementary roles of MSDA and CAVA. Overall, our findings suggest that viewer-centered video emotion analysis should consider not only what a video depicts, but also how its sensory dynamics unfold and how individual modalities shape the resulting affective response.

\clearpage
\bibliography{custom}

\clearpage
\appendix

\setcounter{figure}{0}
\renewcommand{\thefigure}{S\arabic{figure}}
\setcounter{table}{0}
\renewcommand{\thetable}{S\arabic{table}}

\twocolumn[{
\begin{center}
{\large\bfseries PI-MLLM: Psychophysics-Inspired Multimodal Large Language
Model for Oddly Satisfying Video Emotion Analysis\par}
\vspace{0.25em}
{\Large\bfseries Supplementary Material\par}
\end{center}
\vspace{0.5em}
}]

\section{OSVD Details}
\label{sec:dataset_details}

\paragraph{Source.}
OSVD was collected from YouTube videos posted within the five years preceding
collection, retrieved with queries such as ``stress relief,''
``oddly satisfying,'' and related keywords.

\subsection{Modality-Isolated Annotation Prompts}
\label{subsec:prompts}

Molmo2, Music Flamingo, and Claude Sonnet 4.6 produce the video-only,
audio-only, and fused annotations, respectively.  The prompts used in the
annotation pipeline are shown below.  In every prompt, \emph{emotion} must be
exactly one of the 27 viewer-emotion labels defined in the main paper.

\paragraph{Video-only prompt.}\mbox{}\par
\noindent\fbox{\parbox{0.94\columnwidth}{\small
\emph{You are the visual expert for oddly satisfying video emotion analysis.
You receive only the silent five-second video. Predict the dominant emotion
elicited in a typical viewer, not an emotion expressed by a person or animal.
Never infer sound. Ground the analysis in: (1) scene, objects, and materials;
(2) motion speed, repetition, smoothness, regularity, and predictability;
(3) visible state change, deformation, or fragmentation; and (4) color and
lighting. Select exactly one label from the 27-label list. Return JSON only:}
\texttt{\{emotion, scene, motion, state\_change, physical\_cues, reason\}}.}}

\paragraph{Audio-only prompt.}\mbox{}\par
\noindent\fbox{\parbox{0.94\columnwidth}{\small
\emph{You are the audio expert for oddly satisfying video emotion analysis.
You receive only the audio track. Predict the dominant emotion elicited in a
typical listener. Never infer visible objects or actions. Ground the analysis
in: (1) timbre, texture, and frequency content; (2) attack, evolution,
repetition, rhythm, and decay; (3) loudness stability and onset/transient
dynamics; and (4) spatial impression, source proximity, and speech presence.
Select exactly one label from the 27-label list. Return JSON only:}
\texttt{\{emotion, sound\_character, temporal\_envelope,
spatial\_impression, physical\_cues, reason\}}.}}

\paragraph{Audio-visual fusion prompt.}\mbox{}\par
\noindent\fbox{\parbox{0.94\columnwidth}{\small
\emph{You receive the independently generated video-only and audio-only
reports. Predict the dominant viewer emotion after experiencing both
modalities. Do not add facts absent from the reports. Cite concrete evidence
from both modalities and identify their relation as visual-dominant,
audio-dominant, synergy, conflict, or complementary. The final label need not
equal either unimodal label. Return JSON only:}
\texttt{\{emotion, visual\_evidence, audio\_evidence, relation, bridge,
reason\}}.}}

\subsection{Full-Dataset Annotation Distributions}
\label{subsec:distributions}

\begin{figure}[!t]
    \centering
    \includegraphics[width=\columnwidth]{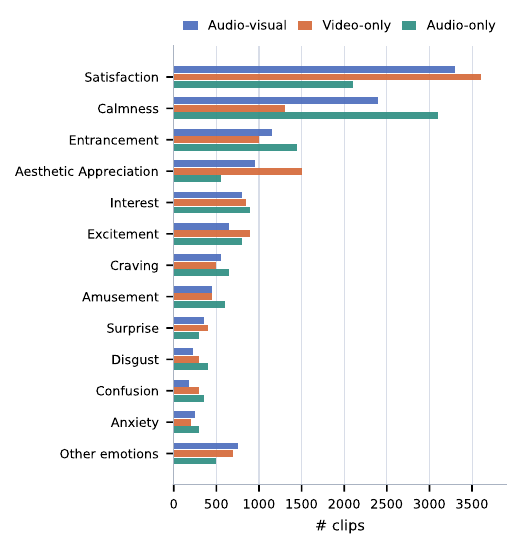}
    \caption{\textbf{Full OSVD three-view label distributions.} Counts are
    computed over all 12,000 clips for audio-visual, video-only, and audio-only
    annotation.}
    \label{fig:annotation_distribution}
\end{figure}

Figure~\ref{fig:annotation_distribution} shows that Satisfaction constitutes
the largest overall share across the three annotation views, while a broad
range of other viewer-emotion categories remains represented.  The
modality-specific distributions are not identical: Satisfaction dominates the
video-only annotations, whereas the audio-only distribution shifts toward
Calmness, showing that isolated acoustic cues can induce an affective prior
different from that of visual dynamics.

\FloatBarrier

\begin{figure*}[!t]
    \centering
    \includegraphics[width=\textwidth,trim=0bp 150bp 250bp 0bp,clip]{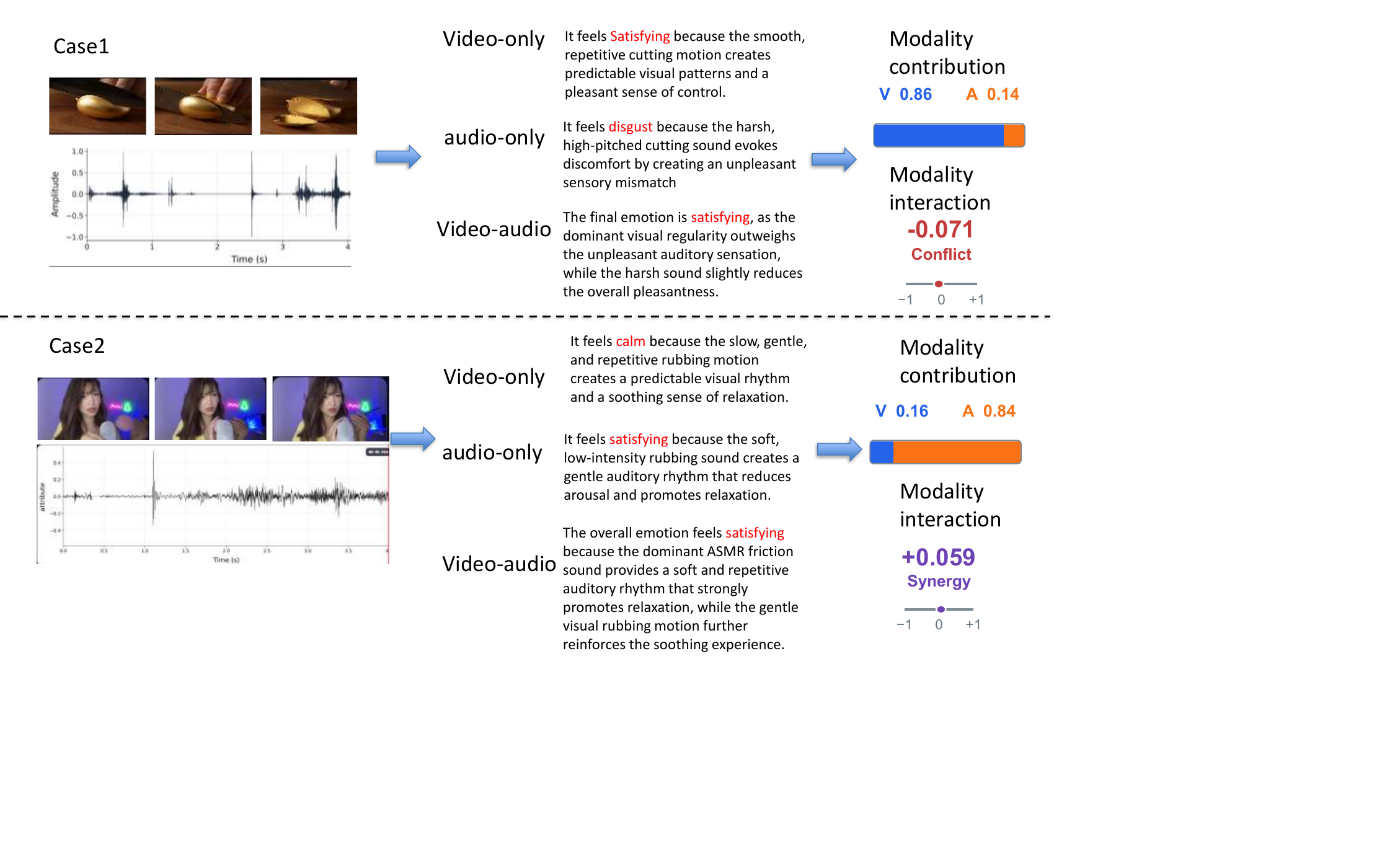}
    \caption{\textbf{Two PI-MLLM cases.} Case 1 is visually dominant
    ($V{=}0.86$, $A{=}0.14$) with negative interaction ($-0.071$, conflict);
    Case 2 is audio dominant ($V{=}0.16$, $A{=}0.84$) with positive
    interaction ($+0.059$, synergy).}
    \label{fig:case_study}
\end{figure*}

\subsection{Annotation Reliability Validation}
\label{subsec:human_agreement}

The main paper reports only the mean Cohen's $\kappa$ for each modality.
Table~\ref{tab:human_full} gives the complete per-annotator results for five
human annotators against the corresponding model-generated labels on the same
100 clips.  The closely clustered scores and the standard deviation of $0.03$
for every view indicate stable agreement patterns across annotators, while the
$\kappa$ range of $0.63$--$0.76$ demonstrates consistently high agreement
between individual annotators and the automatic labels.

\begin{table}[!h]
\centering
\caption{\textbf{Human--model Cohen's $\kappa$} ($N{=}100$ clips per
annotator and view).}
\label{tab:human_full}
\small
\setlength{\tabcolsep}{4.2pt}
\renewcommand{\arraystretch}{1.08}
\begin{tabular}{@{}lccc@{}}
\toprule
Annotator & Video-only & Audio-only & Audio-visual \\
\midrule
Annotator 1 & 0.66 & 0.63 & 0.67 \\
Annotator 2 & 0.71 & 0.69 & 0.73 \\
Annotator 3 & 0.68 & 0.65 & 0.69 \\
Annotator 4 & 0.70 & 0.67 & 0.71 \\
Annotator 5 & 0.74 & 0.72 & 0.76 \\
\midrule
Mean $\pm$ SD & $0.70{\pm}0.03$ & $0.67{\pm}0.03$ & $0.71{\pm}0.03$ \\
\bottomrule
\end{tabular}
\end{table}

\section{Definition of the Empty Coalition in CAVA}
\label{sec:empty_coalition}

Coalition-Based Audio-Visual Attribution (CAVA) represents no video and no
audio by a modality-independent zero vector:
\begin{equation}
\begin{aligned}
z_{\emptyset}&=e_{\emptyset}=\mathbf{0}, &
z_{\{V\}}&=P_V(\bar F_V),\\
z_{\{A\}}&=P_A(\bar F_A), &
z_{\{V,A\}}&=P_V(\bar F_V)+P_A(\bar F_A),
\end{aligned}
\end{equation}
where $P_V$ and $P_A$ project pooled modality features into the shared space.
With the shared classifier,
$f_{\psi}(\emptyset)=W_{\psi}\mathbf{0}+b_{\psi}=b_{\psi}$.
Thus, $f(\emptyset)$ is the classifier-bias baseline used by the Shapley
equations.  It contains no encoded black video, silent waveform, text token,
or pseudo emotion label.

\section{Case Study}
\label{sec:case_study}

The same checkpoint is evaluated under video-only, audio-only, and
audio-visual input.  Figure~\ref{fig:case_study} reports the generated labels
and explanations, normalized Shapley contributions, and signed interaction.
Case 1 shows that regular visual cutting can dominate an unpleasant sound;
Case 2 shows that an ASMR-like sound can dominate while the gentle visual
rhythm reinforces it.

\section{Ablation of Physical-Feature Annotation}
\label{sec:physical_annotation}

The semantic-only target retains object, scene, and event descriptions; the
semantic+physical target additionally retains the visual and acoustic fields
listed in the prompts above.  All other settings are fixed.

\begin{table}[!h]
\centering
\caption{\textbf{Effect of physical-feature annotation.} ACC is 27-class
top-1 accuracy; RS is explanation cosine similarity.}
\label{tab:physical_annotation}
\resizebox{\columnwidth}{!}{%
\begin{tabular}{llcc}
\toprule
Training annotation & Method & ACC (\%) & RS \\
\midrule
Semantic only & Qwen3-Omni + AR fine-tuning & 33.2 & 0.574 \\
Semantic + physical & Qwen3-Omni + AR fine-tuning & 35.1 & 0.636 \\
Semantic only & Qwen3-Omni + PI-MLLM & 38.9 & 0.602 \\
Semantic + physical & Qwen3-Omni + PI-MLLM & \textbf{43.2} & \textbf{0.661} \\
\bottomrule
\end{tabular}}
\end{table}

Table~\ref{tab:physical_annotation} shows that adding physical-feature
annotations improves both training approaches over their semantic-only
counterparts: AR fine-tuning rises from $33.2/0.574$ to $35.1/0.636$ in
ACC/RS, while PI-MLLM rises from $38.9/0.602$ to $43.2/0.661$.  Under the
matched semantic+physical setting, AR fine-tuning still trails PI-MLLM by
$8.1$ ACC points and $0.025$ RS.  Together with PI-MLLM's larger ACC gain from
physical annotations ($4.3$ versus $1.9$ points), these results indicate that
the proposed method effectively learns and exploits the importance of physical
sensory features for viewer-emotion analysis.

\section{Experimental Hyperparameter Design}
\label{sec:implementation_details}

The experimental hyperparameter settings are shown in
Table~\ref{tab:hyperparameters}.

\begin{table}[!h]
\centering
\caption{\textbf{PI-MLLM experimental hyperparameter settings.}}
\label{tab:hyperparameters}
\scriptsize
\setlength{\tabcolsep}{2.5pt}
\renewcommand{\arraystretch}{1.02}
\resizebox{\columnwidth}{!}{%
\begin{tabular}{@{}llll@{}}
\toprule
Hyperparameter & Value & Hyperparameter & Value \\
\midrule
Backbone & Qwen3-Omni-30B-A3B-Thinking & Precision & bf16 \\
Frozen encoders & Visual and audio & LoRA target & Decoder q/k/v/o \\
LoRA rank / alpha / dropout & $64/128/0.05$ & Optimizer / decay & AdamW / $0$ \\
Learning rate / schedule & $10^{-4}$ / cosine & Warmup / grad.\ clip & $10\%/1.0$ \\
Loader batch / accumulation & $8/1$ & Epochs / seed & $5/42$ \\
Video rate / pixel range & $1$ fps / $16{,}384$--$50{,}176$ & Temporal steps $T$ & $10$ \\
$\lambda_1,\lambda_2$ (VMD, AED) & $0.1,0.1$ & $\lambda_3,\lambda_4$ (CLS, CM) & $0.1,0.1$ \\
CAVA classes / hidden / dropout & $27/512/0.1$ & & \\
\bottomrule
\end{tabular}}
\end{table}

\end{document}